\pdfoutput=1

\documentclass{article}

\usepackage[preprint,nonatbib]{neurips_2024}

\usepackage[utf8]{inputenc} 
\usepackage[T1]{fontenc}    
\usepackage{hyperref}       
\usepackage{url}            
\usepackage{booktabs}       
\usepackage{amsfonts}       
\usepackage{nicefrac}       
\usepackage{microtype}      
\usepackage{xcolor}         

\usepackage{caption}
\usepackage{subcaption}
\usepackage{graphicx}
\usepackage{wrapfig}
\usepackage{amsmath}

\usepackage{multirow}

\newcommand{\OM}{DreamGaussian4D}
\newcommand{\OMS}{DG4D}

\title{DreamGaussian4D: Generative 4D Gaussian Splatting}

\author{%
    Jiawei Ren\thanks{Equal contribution
    }$\hspace{0.38em}^{1}$
    \quad
    Liang Pan$^{*2}$
    \quad
    Jiaxiang Tang$^{1,3}$
    \quad
    Chi Zhang$^{1}$
    \quad
    Ang Cao$^{4}$
    \\
    \textbf{Gang Zeng}$^{3}$
    \quad
    \textbf{Ziwei Liu}$^{1}$  \vspace{0.2cm}\\
  $^1$ S-Lab, Nanyang Technological University \quad
  $^2$ Shanghai AI Laboratory\\
  $^3$ Peking University \quad
  $^4$ University of Michigan
}

\begin{document}

\maketitle

\begin{center}
    \centering
    \vspace{-6mm}
    \includegraphics[width=1\textwidth]{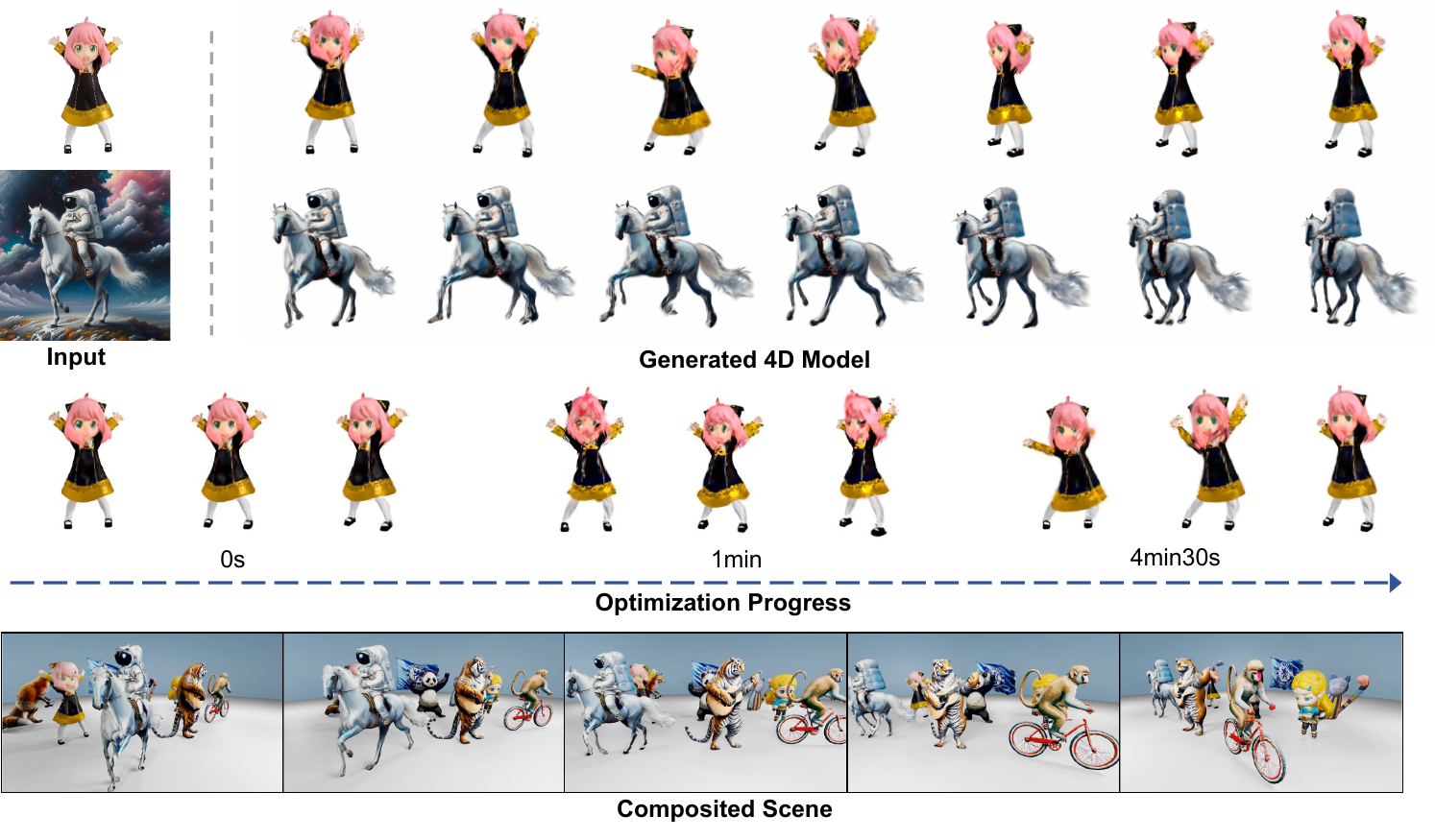}
    \vspace*{-.8cm}
    \captionof{figure}{\textbf{DreamGaussian4D} generates 4D contents in minutes by leveraging 4D Gaussian Splatting. Exported meshes can be efficiently composited and rendered in 3D engines (\emph{e.g.}, Blender or Unreal Engine).}
\label{fig:teaser}
\end{center}

\begin{abstract}
4D content generation has achieved remarkable progress recently. However, existing methods suffer from long optimization times, a lack of motion controllability, and a low quality of details. In this paper, we introduce \textbf{DreamGaussian4D} (\textbf{\OMS{}}), an efficient 4D generation framework that builds on Gaussian Splatting (GS). 
Our key insight is that combining explicit modeling of spatial transformations with static GS makes an efficient and powerful representation for 4D generation.
Moreover, video generation methods have the potential to offer valuable spatial-temporal priors, enhancing the high-quality 4D generation.
Specifically, we propose an integral framework with two major modules: \textbf{1)} Image-to-4D GS - we initially generate static GS with DreamGaussianHD, followed by HexPlane-based dynamic generation with Gaussian deformation; 
and \textbf{2)} Video-to-Video Texture Refinement - we refine the generated UV-space texture maps and meanwhile enhance their temporal consistency by utilizing a pre-trained image-to-video diffusion model.
Notably, \OMS{} reduces the optimization time from several hours to just a few minutes, allows the generated 3D motion to be visually controlled, and produces animated meshes that can be realistically rendered in 3D engines. 
\end{abstract}

\section{Introduction}\label{sect:intro}
Remarkable progress has been witnessed in generative models, demonstrating significant recent advancements and innovations in generating diverse digital content, such as 2D images~\cite{rombach2022high,sheynin2022knn}, videos~\cite{wang2023lavie,blattmann2023stable}, and 3D scenes~\cite{jun2023shap,hong2023lrm,tang2023dreamgaussian}.
While a few recent research works~\cite{singer2023text,jiang2023consistent4d,zhao2023animate124,bahmani20234d} have been devoted to dynamic 3D (\emph{i.e.,} 4D) generation, achieving consistency and high quality in the generation of 4D scenes is far from being fully resolved.

4D scenes are often represented by dynamic Neural Radiance Fields (NeRF), which are expected to show consistent appearance, geometry, and motions from arbitrary viewpoints.
By combining the benefits of video and 3D generative models, MAV3D~\cite{singer2023text} achieves text-to-4D generation by distilling text-to-video diffusion models on a HexPlane~\cite{cao2023hexplane}.
Consistent4D~\cite{jiang2023consistent4d} introduces a video-to-4D framework to optimize a Cascaded DyNeRF for 4D generation from a statically captured static video. 
With multiple diffusion priors, Animate124~\cite{zhao2023animate124} could animate a single in-the-wild image into 3D videos through textual motion descriptions.
Using a hybrid SDS, 4D-fy~\cite{bahmani20234d} achieves compelling text-to-4D generation based on multiple pre-trained diffusion models.
However, all the aforementioned methods ~\cite{singer2023text,jiang2023consistent4d,zhao2023animate124,bahmani20234d} need several hours to generate a single 4D NeRF, which limits their application potential.
Furthermore, it is usually challenging to effectively control their generated motions.
This dissatisfaction comes from several factors. 
First, the underlying implicit 4D representations of the aforementioned methods are not efficient enough, suffering from slow rendering speed and less regularized motions.
Second, the stochastic nature of video SDS adds difficulties to convergence and introduces instability and artifacts to the final results.

\begin{wraptable}{r}{0.5\textwidth}
\setlength{\tabcolsep}{3.pt}
\vspace{-4mm}
\caption{\textbf{4D Generation in Several Minutes.} \OMS{} generates promising 4D assets with significant improvement in convergence speed (iteration numbers) and rendering speed (seconds needed for each iteration).}
\vspace{-4mm}
\fontsize{8}{9}\selectfont
\begin{center}
\begin{tabular}{lcc}
\toprule
Method & Time & Iterations \\
\midrule 
MAV3D~\cite{singer2023text} & 6.5 hr  & 12k\\
Animate124~\cite{zhao2023animate124} & - & 20k\\
Consistent4D~\cite{jiang2023consistent4d} & 2.5 hr & 10k\\
4D-fy~\cite{bahmani20234d} & 23 hr & 120k \\
Dream-in-4D~\cite{zheng2023unified} & 10.5 hr & 20k \\
AYG~\cite{ling2023align} & - & 20k\\
\midrule
Ours (Image-to-4D GS Generation) & \textbf{6.5 mins} & \textbf{0.7k}\\
    \;\; + Video-to-Video Texture Refinement & 10 mins & 0.75k \\

\bottomrule
\label{tab:speed}
\vspace{-0.8cm}
\end{tabular}
\end{center}
\end{wraptable}

In this work, we introduce \textbf{DreamGaussian4D} (\textbf{\OMS{}}), which can efficiently generate dynamic scenes from a single image (or a video sequence) in just a few minutes. 
Our core idea is to leverage \emph{explicit 4D representations} and \emph{video-driven optimization} to accelerate the optimization process for 4D generation. 
For 4D representations, we employ \emph{3D Gaussian Splatting} (GS)~\cite{kerbl20233d} to explicitly represent the static 3D scene and \emph{HexPlane} to describe dynamic displacement maps. 
First, we introduce a new training recipe and propose the image-to-3D framework DreamGaussianHD to initialize a static 3D GS, effectively alleviating the under-optimization problem in DreamGaussian \cite{tang2023dreamgaussian}.
We then optimize a HexPlane to introduce realistic motions into the static 3D GS by predicting Gaussian deformations at each timestamp.

Unlike the commonly used score distillation from video diffusion models, our model learns the motion from a driving video.
In practice, users can select a desired driving video, which can be efficiently generated (typically in about one minute) using an image-to-video generation model (\emph{e.g.,} SVD~\cite{blattmann2023stable} or Sora~\cite{sora_website}). 
This approach allows flexible selection and control of the expected motions in 4D generation while maintaining a high level of diversity in motion generation.
Subsequently, the generated 4D GS could be exported into an animated mesh sequence.
To further improve the temporal coherence of the generated sequence, we can optionally refine the per-frame texture map by using a Video-to-Video optimization pipeline with an image-to-video diffusion model.
In total, generating a 14-frame 4D animation takes approximately 6.5 minutes or 10 minutes with the optional texture refinement (see Table~\ref{tab:speed}).
The generated 4D content by \OMS{} can be visualized in Fig.~\ref{fig:teaser}.

In summary, our contributions are as follows:
\textbf{(1) Principled Image-to-4D Generation Framework:} We propose a principled image-to-4D generation framework that employs a synergy of image-conditioned 3D-generation and video-generation models. 
This allows direct control and selection of the expected 3D content and its motions, enabling high-quality and diverse 4D generation.
\textbf{(2) Explicit 4D Representation:} By utilizing Gaussian Splatting and HexPlane, the proposed \OMS{} explicitly represents 4D scenes with 3D GS and their deformations at different timestamps.
The explicit modeling of spatial transformation significantly reduces the 4D generation time from several hours to just a few minutes.
\textbf{(3) Video-to-Video Texture Refinement:} We introduce a Video-to-Video texture refinement strategy that further enhances the quality of exported animated meshes by maintaining temporal consistency, making the framework more friendly to deploy in a real-world setting.
\textbf{(4) Superior Performance:} Experimental results show that \OMS{} can effectively generate diverse 4D content with higher quality and shorter optimization time than existing methods.

\section{Related works}\label{sect:related_works}

\noindent\textbf{Image-to-3D Generation.}
Image-to-3D generation could be regarded as a conditional generation task that creates 3D assets from a single reference image, often employing advanced techniques such as diffusion models~\cite{ho2020denoising}. 
Point-E~\cite{nichol2022point} and Shap-E~\cite{jun2023shap} are trained to generate 3D point clouds or Neural Radiance Fields (NeRF)~\cite{mildenhall2020nerf} based on image features, but their quality is limited by spatial resolution and the availability of high-quality 3D datasets.
Some methods~\cite{melas2023realfusion,tang2023make} leverage powerful 2D diffusion models~\cite{rombach2022high,liu2023zero,deitke2023objaversexl} and adapt them to 3D using score distillation sampling (SDS)~\cite{poole2022dreamfusion}. 
For instance, Magic123~\cite{qian2023magic123} integrates both image and text inputs to distill high-quality 3D models via NeRF, while DreamGaussian~\cite{tang2023dreamgaussian} reduces optimization times with Gaussian splatting~\cite{kerbl20233d}.
Alternatively, the challenge can be approached as a single-view 3D reconstruction task, where various works~\cite{xu2019disn,chen2019learning,chen2020bsp,trevithick2021grf,duggal2022topologically,szymanowicz2023splatter} employ auto-encoder structures to learn 3D priors for this complex problem, typically constrained to specific categories of synthetic objects~\cite{chang2015shapenet}. 
Recent developments include One-2-3-45~\cite{liu2023one,liu2023one2345++} which uses 2D diffusion models~\cite{liu2023zero,shi2023zero123plus} to generate multi-view images for training an efficient reconstruction model, and LRM~\cite{hong2023lrm} which utilizes a transformer-based architecture to enhance scale on large datasets~\cite{deitke2023objaverse,yu2023mvimgnet} by directly regressing a triplane-based NeRF.

\noindent\textbf{4D Representations.}
Significant advancements have been made in the representation of dynamic 3D scenes~(4D scenes). 
Research has approached 4D scene representation either as a function of spatial coordinates $x, y, z$ with an additional time dimension $t$ or latent codes~\cite{xian2021space, gao2021dynamic, li2022neural, li2021neural}, or by modeling 4D scenes through deformation fields combined with static canonical 3D scenes~\cite{pumarola2021d, park2021nerfies, park2021hypernerf, du2021neural, tretschk2021non, yuan2021star, li2023dynibar}. 
A primary challenge in 4D representations is the extensive computational time, often requiring many hours for a single scene.
To address this, many approaches achieve impressive 4D reconstruction results by utilizing explicit or hybrid representations, including planar decomposition for 4D space-time grids~\cite{cao2023hexplane, fridovich2023k, shao2023tensor4d}, hash representations~\cite{turki2023suds}, and other structures~\cite{fang2022fast, abou2024particlenerf, guan2022neurofluid}. 
Recently, Gaussian Splatting~\cite{kerbl20233d} has gained attention for its balance of speed and quality in reconstructions. 
The extension of static Gaussian Splatting into dynamic contexts, such as Dynamic 3D Gaussians~\cite{luiten2023dynamic} and 4D Gaussian Splatting~\cite{wu20234d,yang2023deformable}, utilizes deformation networks to predict time-dependent adjustments, offering a promising direction in 4D scene representation.

\noindent\textbf{4D Generation.}
The objective of 4D generation is to create dynamic 3D scenes, applicable across various graphics fields such as animation, gaming, and virtual reality. 
Current methodologies employ text-to-video diffusion models to distill 4D content~\cite{singer2023text}, notably using advanced 4D representations like Hexplane~\cite{cao2023hexplane} or K-plane~\cite{fridovich2023k}. 
These models synthesize camera trajectories and calculate SDS on the rendered videos. 
Recent advancements have aimed to enhance photorealism by integrating multiple diffusion priors, providing a more robust supervision signal~\cite{bahmani20234d, zheng2023unified}. 
However, the excessive optimization time and computational demands limit practical deployment. 
Additionally, the 3D content often lacks motion diversity and control. 
Recent initiatives attempt to generate 4D models from single images, yet they still face challenges related to lengthy optimization and insufficient motion control. 
Notably, Consistent4D~\cite{jiang2023consistent4d} offers a novel approach by deriving 4D models from static input videos, closely aligning with our work. 
Our study differs by focusing on image-conditioned video generation, enabling varied motions using the same static model. 
Concurrently, research like AYG~\cite{ling2023align} utilizes Gaussian Splatting for high-fidelity 4D generation, but our method achieves comparable results with fewer than 5$\%$ of the optimization iterations required by these approaches.

\begin{figure*}[t!]
\begin{center}
\includegraphics[width=\textwidth]{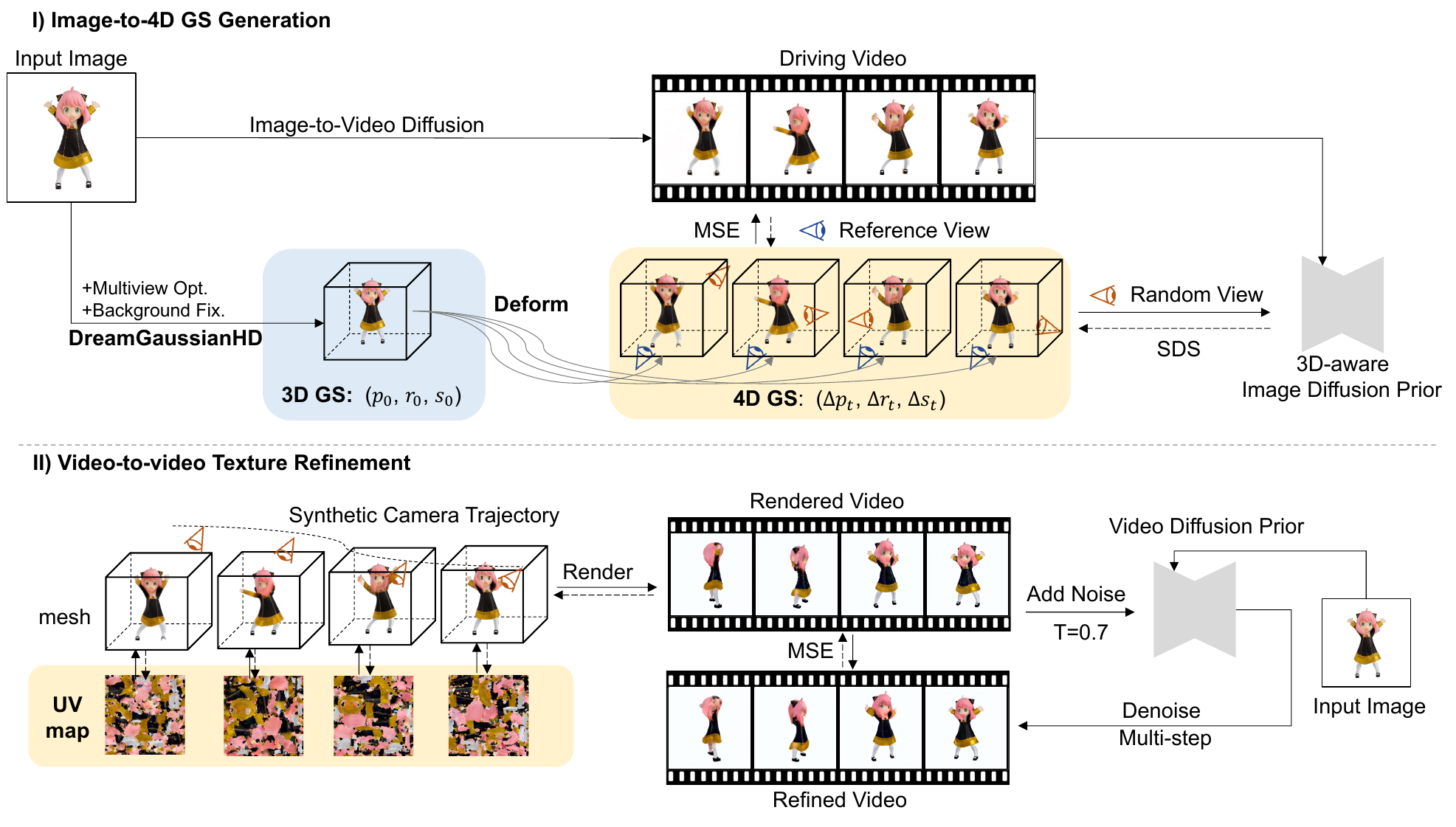}
\vspace{-7mm}
\caption{\textbf{DreamGaussian4D Framework.} We first obtain a static 3D GS model using {DreamGaussianHD} and a driving video with an image-to-video diffusion model. We then optimize a deformation network that learns to deform the static 3D GS at different time stamps, supervised by the MSE loss to the driving video and SDS losses. Finally, per-frame meshes can be exported and the texture maps can be refined with a video-to-video pipeline.
}
\label{fig:framework}
\end{center}
\end{figure*}

\section{Our Approach}\label{sect:methods}

As illustrated in Fig.~\ref{fig:framework}, \OMS{} mainly consists of two stages, 1) Image-to-4D GS Generation, and 2) Video-to-Video Texture Refinement.
In the first stage, we represent the dynamic 4D scene using static 3D GS and its deformations. 
From an input image, a static 3D GS is generated using the enhanced method, DreamGaussianHD.  
Subsequently, Gaussian deformations at various timestamps are estimated by optimizing a time-dependent deformation field on the static 3D Gaussians, ensuring that the shape and texture of each deformed frame correspond to every frame of the video generated based on the input image.
This process results in an animated mesh sequence.
In the second stage, our objective is to refine the texture maps of this mesh sequence to enhance temporal consistency.
This is achieved through a Video-to-Video refinement pipeline that utilizes a pre-trained image-to-video diffusion model to improve texture quality.
The entire framework can be completed in approximately 10 minutes, with the first stage requiring about 6.5 minutes and the second stage about 3.5 minutes.

\subsection{Image-to-4D GS Generation}

\subsubsection{DreamGaussianHD for Static Generation} 
\begin{wrapfigure}{r}{0.5\textwidth}
    \vspace{-8mm}
    \begin{center}
       \includegraphics[width=\linewidth]{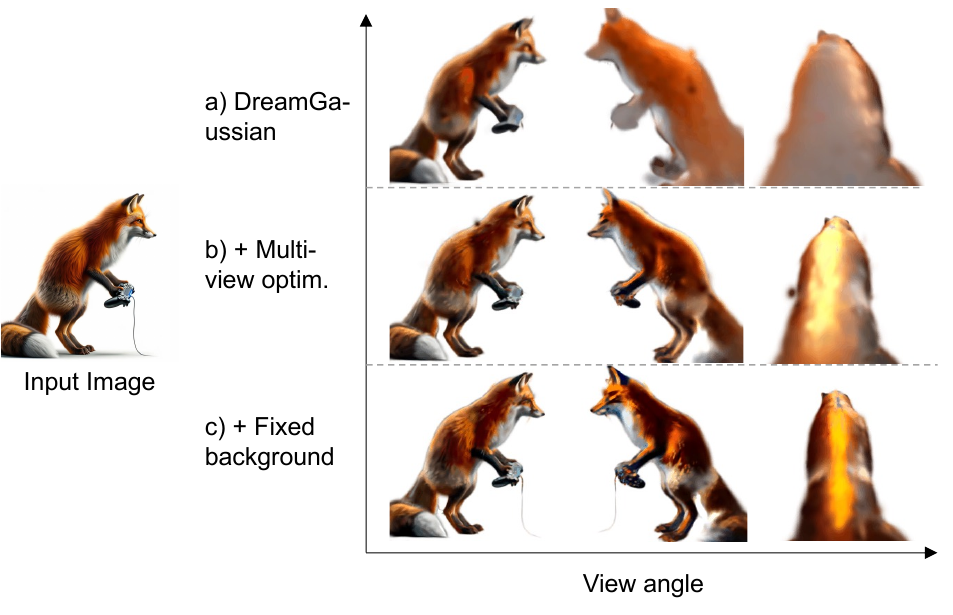}
        \vspace{-6.75mm}
        \caption{\textbf{DreamGaussianHD.} 
        \textit{Multi-view Optimization} significantly improves the texture and geometries. \textit{Fixing Background Color} further enhances the level of detail.}
        \label{fig:dghd}
    \end{center}
    \vspace{-6.5mm}
\end{wrapfigure}

Despite its rapid optimization speed, the original DreamGaussian~\cite{tang2023dreamgaussian} introduces significant blurriness to the unseen areas of static models, as illustrated in Fig.~\ref{fig:dghd}. 
This blurriness adversely affects the subsequent dynamic optimization process. 
Therefore, we first design better implementation practices to reliably enhance the image-to-3D generation quality of DreamGaussian at the cost of a reasonable increase in optimization time. 
We summarize these improved practices as DreamGaussianHD.

\textbf{(1) Multi-view Optimization.}
Apart from the reference view, DreamGaussian typically samples one random view at each optimization iteration for SDS. 
This approach covers only part of the Gaussians and leads to unbalanced optimization and convergence.
As observed in previous works, increasing the number of sampled views (batch size) at each optimization step can significantly mitigate this issue~\cite{poole2022dreamfusion,chen2023fantasia3d}. 
Sampling 16 views, for instance, yields high-quality geometry in the unseen regions of the 3D Gaussians. As a trade-off, this approach incurs an increase in memory usage during SDS computation and lengthens the optimization duration.

\textbf{(2) Fixing Background Color.} 
DreamGaussian uniformly samples the background color from black and white. 
However, most 3D-aware image diffusion models render the training objects with a white background. 
We have observed that renderings with a black background introduce additional noise into the optimization process, ultimately resulting in blurriness. 
By consistently setting the background color to white, we achieve more detailed and refined results in the optimized 3D GS.

\subsubsection{Gaussian Deformation for Dynamic Generation}

\begin{wrapfigure}{r}{0.6\textwidth}
    \vspace{-5.mm}
    \begin{center}
        \includegraphics[width=\linewidth]{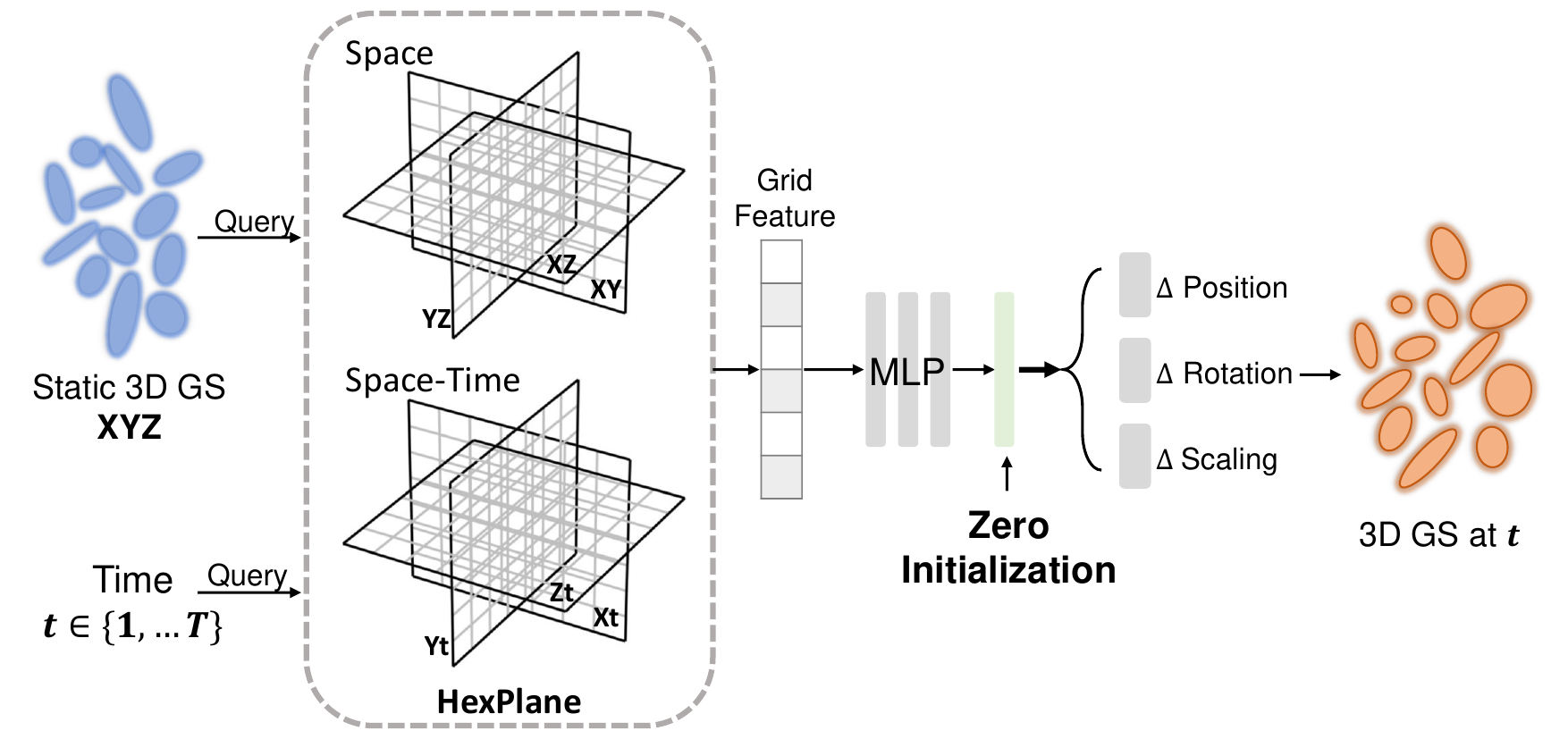}
        \vspace{-5.75mm}
        \caption{\textbf{Deformation Field with HexPlane.} HexPlane provides an explicit mapping for deformation. 
        The resolution provides a regularization to the generated motion. 
        }
        \label{fig:hexp}
    \end{center}
\end{wrapfigure}

Unlike other methods~\cite{bahmani20234d} that perform SDS supervision using a video diffusion model, we propose to use explicit supervision from any video depicting the input image.
This video can be created by artists like those in video-to-4D~\cite{jiang2023consistent4d}, or generated automatically from an image-to-video model.
In this study, we mainly utilize Stable Video Diffusion~\cite{blattmann2023stable}, the current open-source SoTA video generation model, to generate videos from input images:
\begin{align}
     \{I_\mathrm{Ref}\}_{\tau=1}^\mathcal{T} = f_\psi(\epsilon; I_\mathrm{Input}),
\end{align}
where \(I_\mathrm{Input}\) represents the input image, \(\{I_\mathrm{Ref}\}_{\tau=1}^\mathcal{T}\) is the driving video, \(\epsilon\) denotes random noise, and \(f_\psi\) is the image-to-video diffusion model.
Since our method does not rely on the video diffusion model later, users can choose high-quality videos with better temporal consistency and motion generated by different random seeds, which enables better visual controllability and diversity for image-to-4D generation.

\textbf{HexPlane for 4D Representation.}
To further augment the static 3D Gaussians into dynamic ones, we train a HexPlane~\cite{cao2023hexplane} as the 4D representation to predict the position displacement, rotation, and scale of each Gaussian given its location $(x, y, z)$ and timestamp $t$.
As shown in Fig.~\ref{fig:hexp}, HexPlane~\cite{cao2023hexplane} decomposes a 4D field into six feature planes, spanning each pair of coordinate axes. 
Besides the fast speed, this decomposition represents 4D fields as weighted summations of a set of learnable 4D basis functions, which inherently regularizes features of 4D fields and ensures their smoothness (see original paper~\cite{cao2023hexplane} for this discussion). 
In our case,  controlling the spatial and temporal axis resolution of HexPlane allows us to regularize the local spatial rigidness and temporal abruptness of motions, leading to better results. 
Specifically, we extract features from HexPlane representation and regress position displacement, rotation change, and scale changes from an MLP decoder.

\begin{wrapfigure}{r}{0.5\textwidth}
    \vspace{-6mm}
    \begin{center}
       \includegraphics[width=\linewidth]{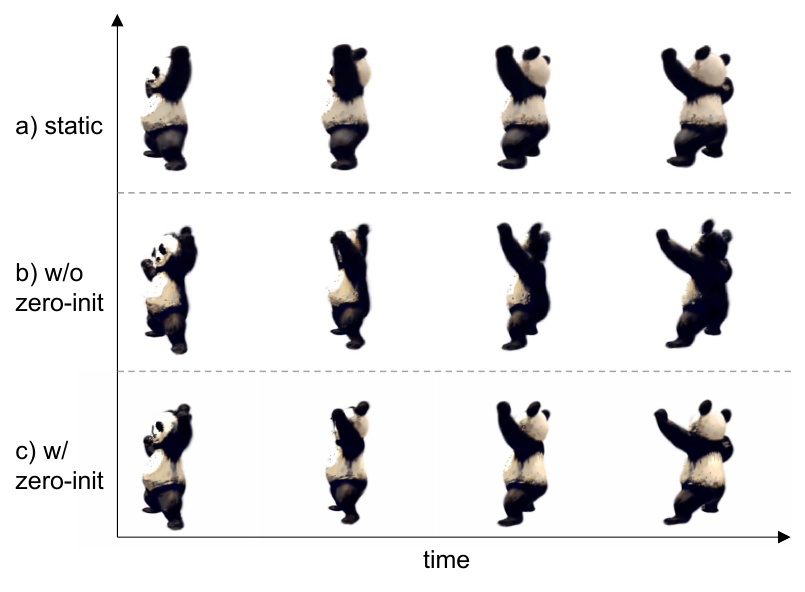}
        \vspace{-7mm}
        \caption{\textbf{Zero Initialization.} 
        Non-zero initialization can induce undesirable random changes, such as turning the back of the panda completely black.
        }
        \label{fig:ablation_zeroinit}
    \end{center}
    \vspace{-3mm}
\end{wrapfigure}

\textbf{Static-to-Dynamic Initialization.}
However, randomly initializing the deformation network can cause a divergence between the dynamic and static models, leading to convergence at a sub-optimal mode. 
As exemplified in Fig.~\ref{fig:ablation_zeroinit}, the back of the panda was black and white in the static stage. 
When initialized differently, the back could turn fully black after dynamic optimization.
To mitigate this, the deformation model should be initialized to predict zero deformation at the start of training. 
Note that if we initialize all the network weights with zero, the weights cannot be updated due to the zero gradient issue.
Therefore, we utilize zero-initialization for the final few linear layers and employ several residual connections to ensure that the weights and inputs of each layer are not simultaneously zero.

\textbf{Deformation Field Optimization.}
We optimize the deformation field given the driving video from the reference view. 
We fix the camera to the reference view, and minimize the Mean Squared Error (MSE) between the rendered image and the driving video frame at each timestamp:
\begin{align}
    \mathcal{L}_\textrm{Ref} = \frac{1}{\mathcal{T}}\sum_{\tau=1}^{\mathcal{T}}||f(\phi(S, \tau),o_\textrm{Ref}) - I^\tau_\textrm{Ref}||^2_2,
\end{align}
where $I^\tau_\textrm{Ref}$ is the $\tau$-th frame in the video, $o_\textrm{Ref}$ is the reference view point and $f$ is the rendering function.
To propagate the motion from the reference view to the whole 3D model, we leverage Zero-1-to-3-XL~\cite{deitke2023objaversexl} to predict the deformation of the unseen part. 
Although image diffusion models only perform per-frame prediction, the temporal consistency can be mostly preserved thanks to the regularization provided by HexPlane.
Similar to the training practice in DreamGaussianHD, multiple views are sampled for each time step.
\begin{align}
    \nabla_{\phi} \mathcal{L}_\textrm{SDS} &=  \mathbb{E}_{t, \tau, \epsilon, o} [(\epsilon_\theta(\hat{I}; t, I^\tau_\textrm{Ref} ,o) - \epsilon)\frac{\partial I}{\partial \phi}],\; \text{and }
    \hat{I} = f(\phi(S, \tau), o),
\end{align}
where $\epsilon$ is a random noise, $\epsilon_\theta$ is the noise predictor of a 3D-aware image diffusion model, and $o$ is a random viewpoint.
Thanks to the static model initialization, we can start the SDS at a lower noise level. 
Specifically, we start SDS with a $T_\mathrm{max} = 0.5$, which is lower than the common practice where $T_\mathrm{max}=0.98$. Optionally, the static 3D GS can be fine-tuned in the optimization, and we freeze the static 3D GS by default

\subsection{Video-to-Video Texture Refinement}
\label{sec:refine}

Following the initial 4D GS generation stage, we obtain a continuous sequence of 3D scenes. 
Mesh extraction for each frame could be conducted similarly to DreamGaussian~\cite{tang2023dreamgaussian}, by performing local density queries and color back-projection. 
The resulting 3D mesh sequences exhibit impressive generative quality, especially in 3D geometries and 4D movements, closely aligning with the 2D animations chosen by the user.
However, the textures in the meshes derived from 3D Gaussians tend to be blurry due to SDS ambiguity.

\begin{wrapfigure}{r}{0.375\textwidth}
    \vspace{-7.5mm}
    \begin{center}
        \includegraphics[width=\linewidth]{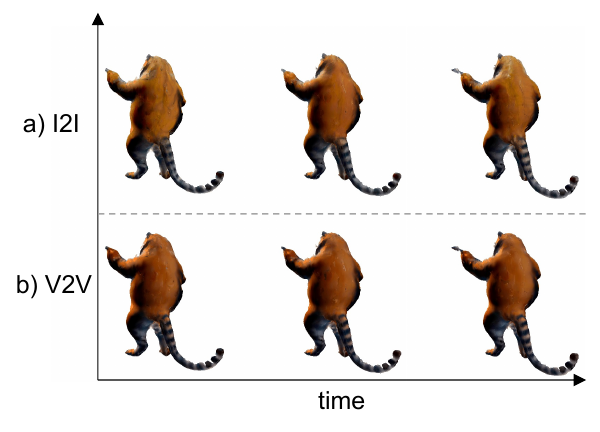}
        \vspace{-7mm}
        \caption{\textbf{Video-to-Video Texture Refinement.} DreamGaussian-like image-to-image (I2I) optimization leads to poor temporal consistency and flickering in adjacent frames. Video-to-Video (V2V) optimization alleviates the issue.}
        \label{fig:ablation_refine}
    \end{center}
    \vspace{-4.mm}
\end{wrapfigure}

To address this, similar to DreamGaussian, a texture refinement module could be introduced, allowing users to fine-tune UV-space textures. 
Unlike the per-frame UV-space refinement in DreamGaussian, which denoises each frame separately and may cause flickering in adjacent frames, 
we employ a Video-to-Video pipeline to enhance the UV-space texture map while maintaining temporal consistency.
This process begins with synthesizing a camera trajectory, where the camera moves at a constant speed along 0 elevations from a randomly chosen horizontal angle. 
We then render the video and introduce noise at level 0.7 to it. 
Finally, an image-to-video diffusion model is utilized to transform this noisy video into a clean, denoised version:
\begin{align}
    \{I_\mathrm{Refined}\}_{\tau=1}^\mathcal{T} = f_\psi(\{\hat{I}\}_{\tau=1}^\mathcal{T} + \epsilon; I_\mathrm{Input}),
\end{align}
where $\epsilon$ is a random noise at the specified level and $\{\hat{I}\}_{\tau=1}^\mathcal{T}$ is the rendered video.
The MSE loss is computed between the two videos:
\begin{align}
    \mathcal{L}_\mathrm{Refine} = ||\{\hat{I}\}_{\tau=1}^\mathcal{T} - \{I_\mathrm{Refined}\}_{\tau=1}^\mathcal{T}||_2^2.
\end{align}
The loss is then back-propagated to improve the texture maps at all time steps.
As shown in Fig.~\ref{fig:ablation_refine}, 
the image-to-image optimization has no temporal consistency restriction since the per-frame refined meshes have individual texture maps.
In contrast, the utilized video-to-video texture refinement provides temporal consistency, which results in smoother temporal changes.

\begin{figure*}[t]
\begin{center}
\includegraphics[width=\textwidth]{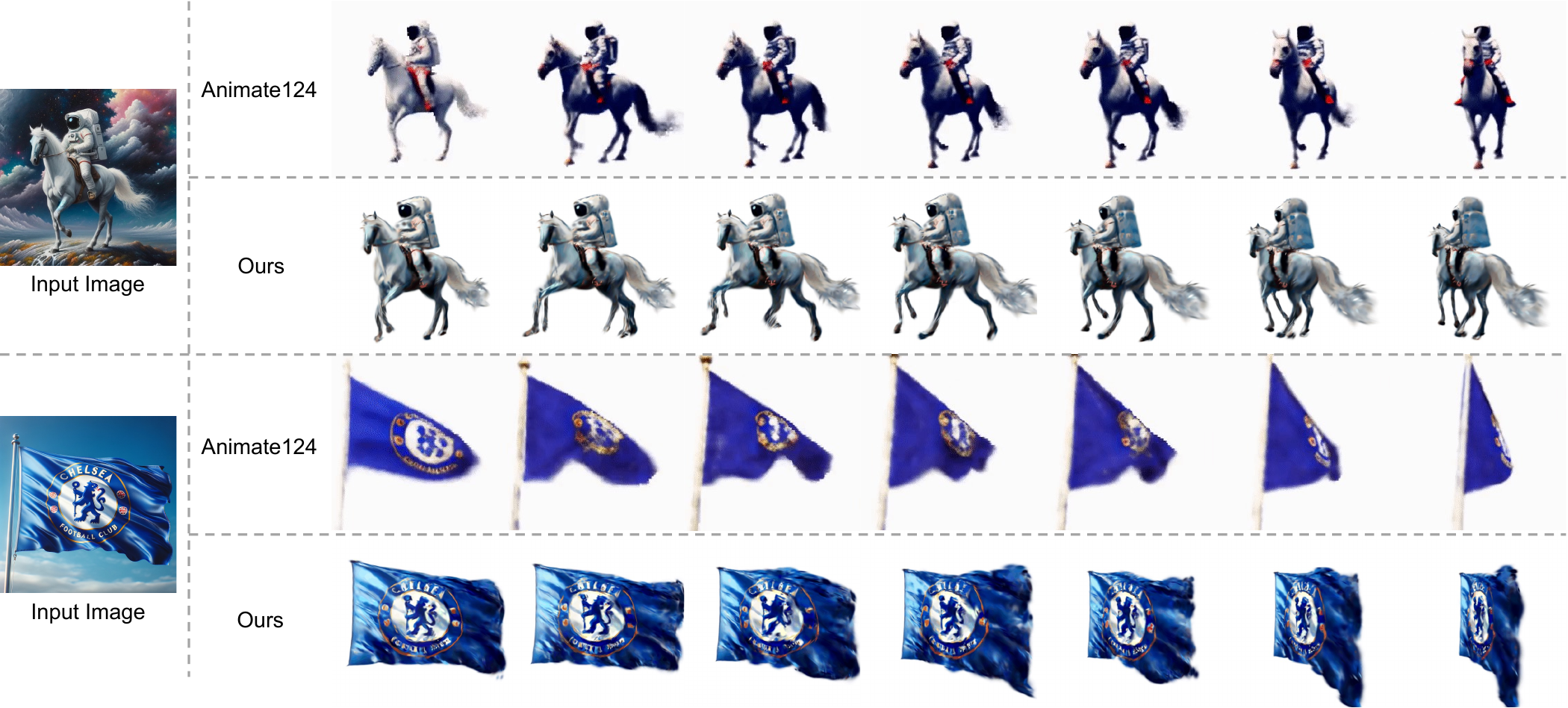}
\vspace{-6mm}
\caption{
\textbf{Qualitative Results for Image-to-4D.} Comparing against Animate124~\cite{zhao2023animate124}, our model achieves better faithfulness to the input image, larger motions, and greater details.
}
\label{fig:comparison}
\end{center}
\end{figure*}

\section{Experiments}\label{sect:experiment}

\noindent\textbf{Implementation Details.}
We run all experiments on a single 80 GB A100 GPU. We implement the \OMS{} framework on the open-source repositories DreamGaussian~\cite{tang2023dreamgaussian} and 4D Gaussian Splatting~\cite{wu20234d}. 
For driving video generation, we use Stable Video Diffusion to generate 14 frames. 
For static optimization, we run 500 iterations with a batch size of 16 for 2 minutes. 
We linearly decay $T_\mathrm{max}$ from 0.98 to 0.02. 
For dynamic representation, we run 200 iterations with batch size 4 for 4.5 minutes, with $T_\mathrm{max}$ linearly decaying from 0.5 to 0.02.
For the optional mesh refinement, we run 50 iterations with a constant $T=0.7$ for 3.5 minutes.

\begin{figure}
    \centering
    \begin{minipage}[t]{0.25\textwidth}
        \setlength{\tabcolsep}{4pt}
        \centering
        \vspace{-37mm}
        \captionof{table}{\textbf{Quantitative Results on Image-to-4D Image Alignment.} $\dagger$: computed on 8 examples available at~\cite{zhao2023animate124}.}
        \vspace{-1.5mm}
        \fontsize{8}{9}\selectfont
        \begin{tabular}{lc}
            \toprule
            Method &CLIP-I$\uparrow$\\  
            \midrule 
            Zero-1-to-3-V  & 0.7925  \\
            RealFusion-V  & 0.8026 \\
            Animate124 & 0.8544 \\
            \midrule
            Ours$^\dagger$ & \textbf{0.9227}\\
            \bottomrule
            \label{tab:quantitative_results}
        \end{tabular}
    \end{minipage}
    \hfill
    \begin{minipage}[t]{0.73\textwidth}
        \raggedright
        \includegraphics[width=1.025\linewidth]{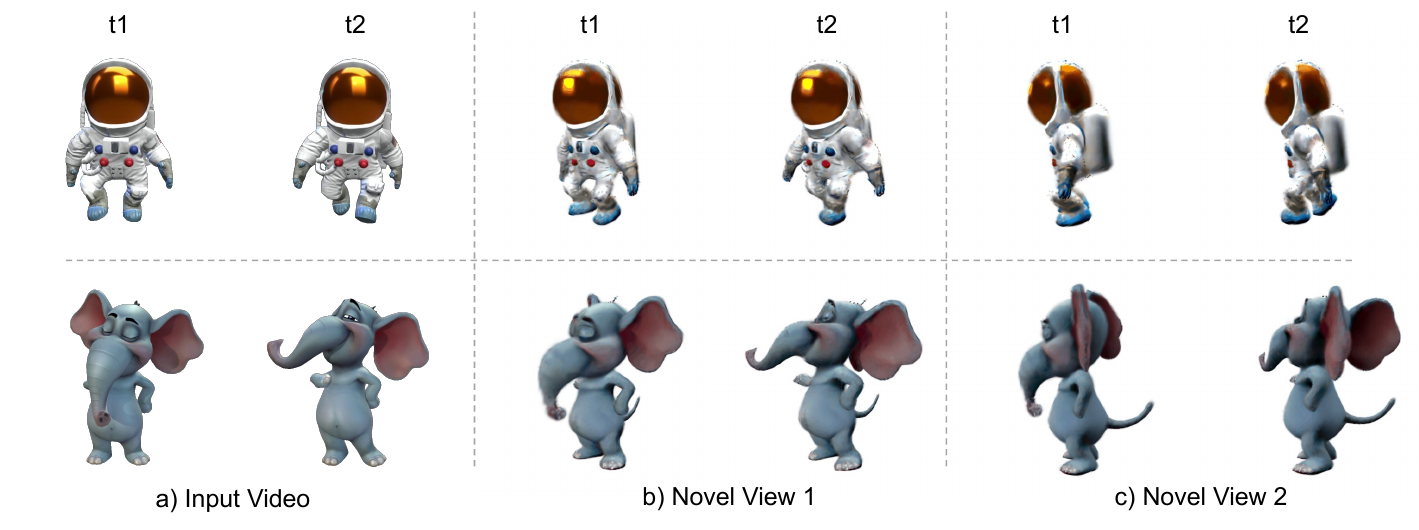}
        \vspace{-6.5mm}
        \caption{\textbf{Qualitative Results for Video-to-4D.} Renders are shown in two novel views at two timesteps.}
        \label{fig:qualitative_v24d}
    \end{minipage}
\end{figure}

\subsection{Comparisons with State-of-the-Art Methods}

\noindent\textbf{Image-to-4D.}
A qualitative image-to-4D comparison is shown in Fig.~\ref{fig:comparison}. 
We collect results from \OMS{} and Animated124 conditioned on 12 images and conduct a user study on 38 participants. 
Participants are asked to select the best one in three evaluation axes: image alignment, 3D appearance, and motion quality. 
The results are presented in Table~\ref{tab:user_study}, where our approach achieves a high win rate across all axes.
We also quantify the image alignment using the metric CLIP-I provided by~\cite{zhao2023animate124}. 
CLIP-I measures the cosine similarity of CLIP image embedding between reference-view renders and the reference image. 
The results are provided in Table~\ref{tab:quantitative_results}, where \OMS{} has a clear advantage. 
Moreover, \OMS{} has a significantly shorter optimization time as shown in Table~\ref{tab:speed}. 
All results are without refinement for \OMS{}.

\noindent\textbf{Video-to-4D.}
Our approach is also compatible with the video-to-4D setting~\cite{jiang2023consistent4d}. 
A qualitative video-to-4D result is provided in Fig.~\ref{fig:qualitative_v24d}. We quantitatively evaluate our approach on a benchmark provided by~\cite{jiang2023consistent4d}. 
Seven 32-frame static-view videos are evaluated, and metrics are computed with their ground-truth novel views. 
We use the first frame of each video for the static 3D generation. 
Two variants of our approach are used: \emph{Ours-Fast} downsamples the video to 16 frames and trains 500 iters with one random view, and \emph{Ours} uses all 32 frames and trains 500 iterations with four random views. 
The results are shown in Table~\ref{tab:consistent4d_benchmark}, where both variants of our approach significantly outperform the baseline using much less time.

\begin{figure}
    \centering
    \begin{minipage}[t]{0.52\textwidth}
        \centering
        \vspace{-69mm}
        \includegraphics[width=\textwidth]{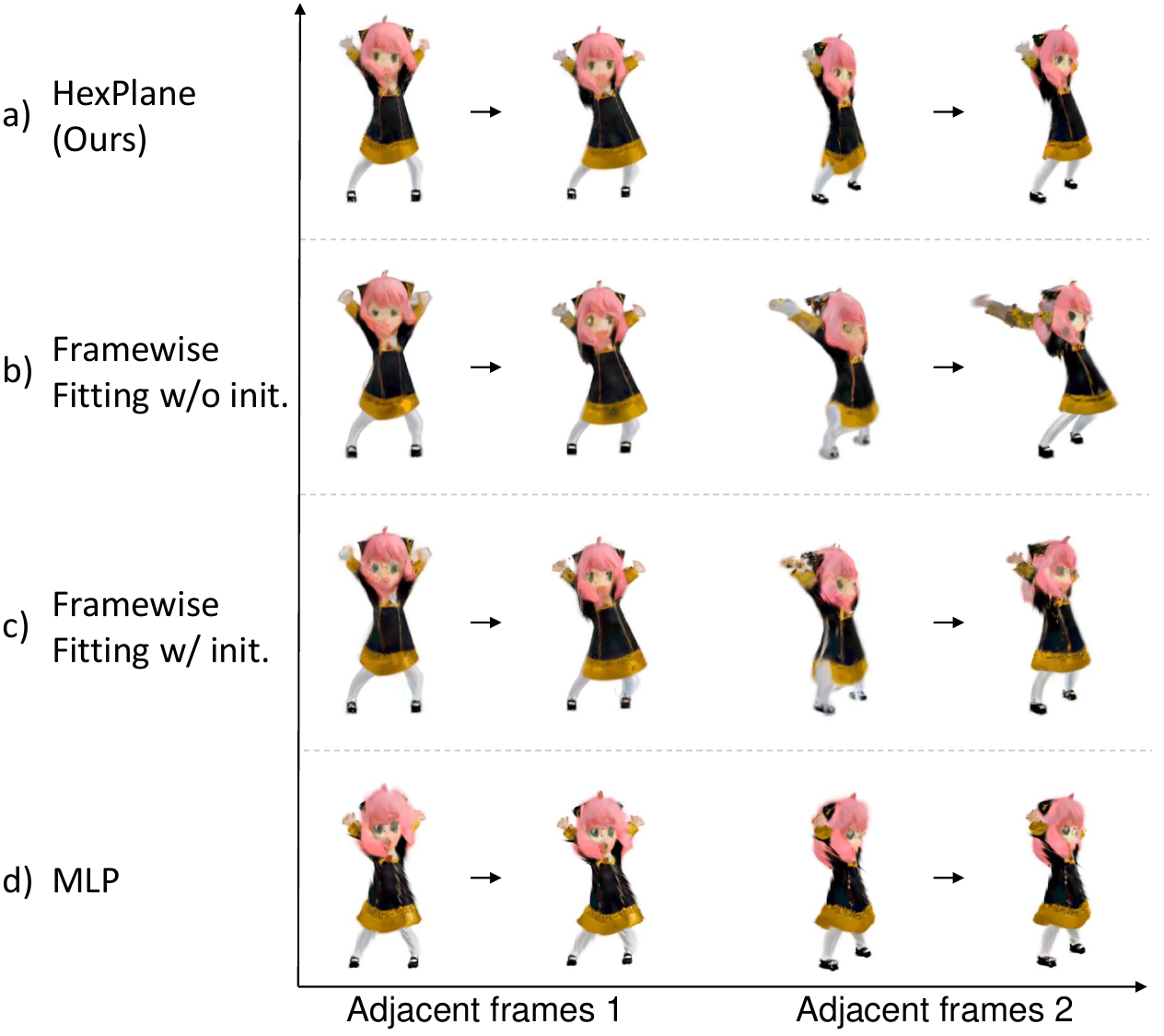}
        \vspace{-2.mm}
        \caption{\textbf{Motion Representation Analysis.}
        HexPlane with static initialization achieves the best generation performance.
        }
        \label{fig:motion_rep}
    \end{minipage}
    \hfill
    \begin{minipage}[t]{0.45\textwidth}
        \centering
        \includegraphics[width=\textwidth]{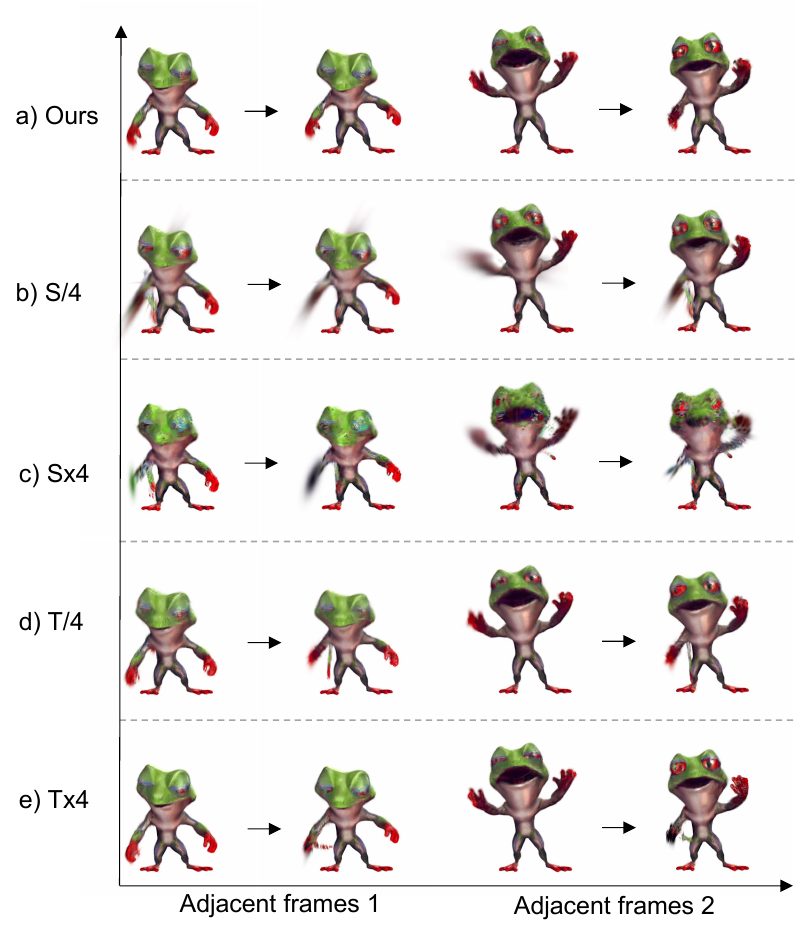}
        \vspace{-7mm}
        \caption{\textbf{HexPlane Resolution Analysis.} 
        \textbf{S} is spatial resolution and \textbf{T} is temporal resolution.}
        \label{fig:ablation_resolution}
    \end{minipage}
\end{figure}

\begin{figure}
    \centering
    \begin{minipage}[t]{0.49\textwidth}
        \centering
        \includegraphics[width=\linewidth]{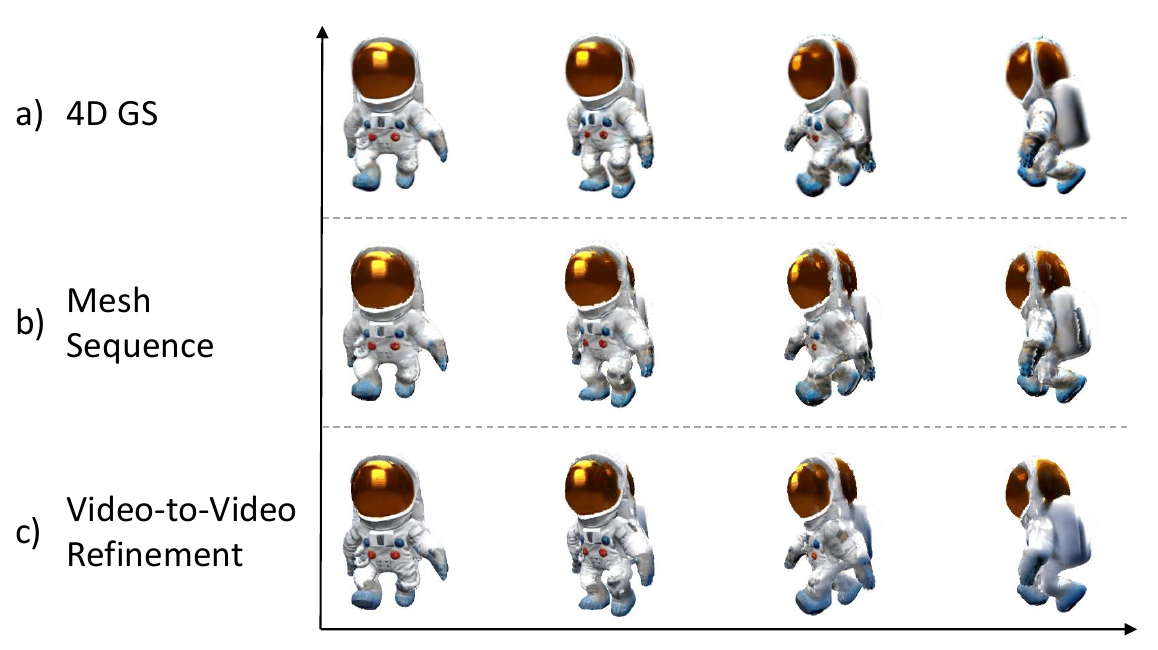}
        \vspace{-5mm}
        \caption{\textbf{Mesh Texture Refinement.} 
        After V2V texture refinement, the texture quality and consistency could be highly improved.
        }
        \label{fig:vv_refinement}
    \end{minipage}
    \hfill
    \begin{minipage}[t]{0.49\textwidth}
        \centering
        \includegraphics[width=\linewidth]{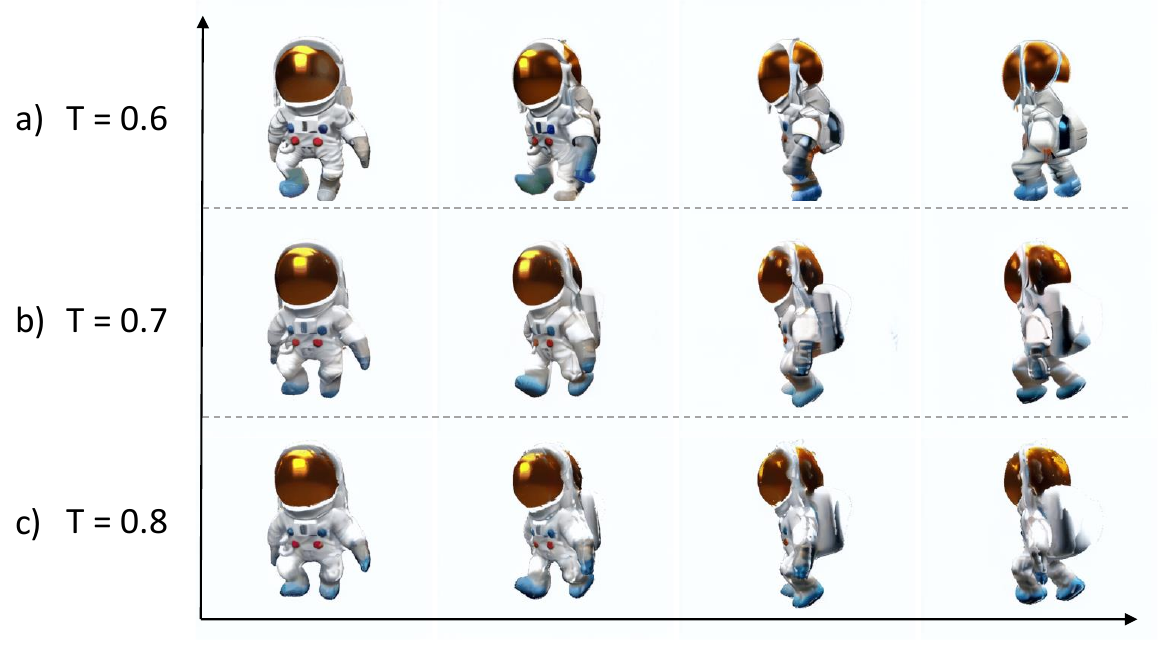}
        \vspace{-5mm}
        \caption{\textbf{Noise Level Analysis.} 
        Different noise levels T used in video SDS refinement.
        }
        \label{fig:noise_level}
    \end{minipage}
\end{figure}

\subsection{Ablation Studies and Analysis}

\textbf{Motion Representation.}
Motion representation is essential for generating 4D content. 
We evaluated various motion representations, including framewise 3DGS with and without static initialization, MLP, and HexPlane.
Employing framewise 3DGS without static initialization effectively conducts 3D generation independently for each frame, often leading to motion blur, resulting in noisy and inaccurate 3D models. 
Conversely, using static initialization enhances the clarity, though some blurriness persists.
Framewise fitting is time-intensive, requiring several minutes per frame, significantly extending the duration for generating sequences (e.g., 14 frames). 
Overall, motion representations generally yield higher quality generation compared to framewise fitting. 
However, using MLP for learning motions frequently produces misalignments and minor local jitter.
Among the evaluated methods, HexPlane consistently delivers superior performance. 
Qualitative comparisons are illustrated in Fig.~\ref{fig:motion_rep}.

\noindent\textbf{HexPlane Resolution.} 
We analyze the effect of HexPlane resolution in Fig.~\ref{fig:ablation_resolution}. 
Empirically, the spatial resolution affects the rigidity of the motion, larger resolution gives more flexibility to the structure change but also challenges the optimization. 
The temporal resolution affects the smoothness of motion, a too-low temporal resolution causes some Gaussian to lag behind the reference motion, and a too-high resolution adds difficulty to optimization.
We use the HexPlane with resolution $32\times32\times32$ (\emph{i.e.,} $H = 32$, and $T = 32$), which yields the best generation results.

\textbf{Mesh Texture Refinement.}
\OMS{} enables users to extract mesh sequences following the acquisition of 4D GS in the initial stage.
While the rendered images from GS typically display impressive outcomes, the textures of the extracted meshes may exhibit noise and artifacts. 
By implementing Video-to-Video (V2V) texture refinement, users can enhance the UV texture quality and consistency across the 3D mesh sequence.
The qualitative improvements are demonstrated in Fig.~\ref{fig:vv_refinement}.

\textbf{Noise Level.}
We also evaluate different noise levels for the video SDS optimization.
The qualitative generation results are visualized in Fig.~\ref{fig:noise_level}.
The refined texture maps could be over-smoothed if using a small noise level (\emph{e.g.}, T = 0.6).
Conversely, the refined texture maps could be too noisy if using a large noise level (\emph{e.g.}, T = 0.8).
We also evaluate the setting, annealing T from 0.7 to 0.95, which often leads to noisy results.
Hence, we choose to fix T = 0.7 in our experiments.

\subsection{Advantages and Applications}
\begin{wrapfigure}{r}{0.5\textwidth}
    \vspace{-7.5mm}
    \begin{center}
        \includegraphics[width=\linewidth]{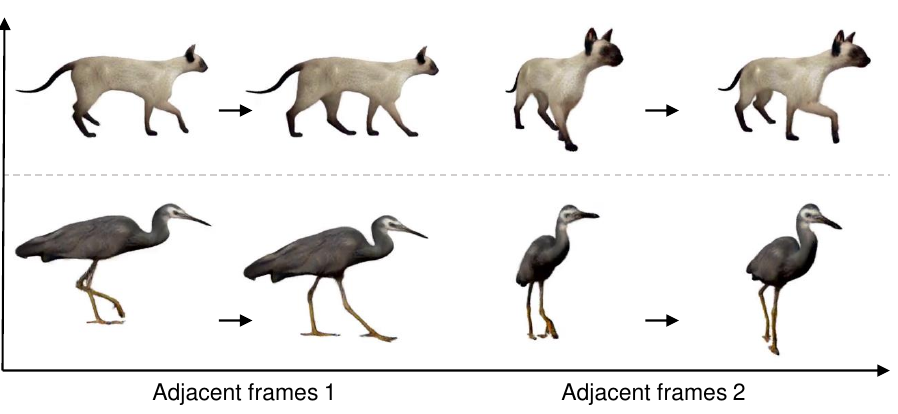}
        \vspace{-6.mm}
        \caption{\textbf{4D Generation based on LGM~\cite{tang2024lgm}.} 
        }
        \label{fig:lgm4d}
    \end{center}
    \vspace{-6.5mm}
\end{wrapfigure}

\begin{figure}
    \centering
    \begin{minipage}[t]{0.46\textwidth}
        \vspace{-2mm}
        \setlength{\tabcolsep}{3.6pt}
        \captionof{table}{\textbf{Quantitative Results for Video-to-4D.} Best is bolded and second best is underlined.}
        \vspace{-4mm}
        \fontsize{8}{9}\selectfont
        \begin{center}
            \begin{tabular}{lcccc}
                \toprule
                Method & LPIPS$\downarrow$ & CLIP$\uparrow$ & FVD$\downarrow$ & Time$\downarrow$\\  
                \midrule 
                D-NeRF  & 0.51 & 0.68 & 2327.83 & -\\
                K-planes  & 0.38 &0.72 & 2295.68 & -\\
                Zero123 & 0.15 & 0.90 & 1571.60 & -\\
                Consistent4D & 0.16 & 0.87 & 1133.44 & 2 hrs\\
                \midrule
                Ours-Fast & \underline{0.13} & \underline{0.91}  &	\underline{775.90} & \textbf{5 mins}\\
                Ours & \textbf{0.12} & \textbf{0.92} &\textbf{729.74} & \underline{15 mins}\\
                \bottomrule
                \label{tab:consistent4d_benchmark}
            \end{tabular}
            \caption{\textbf{User Study. 
            }}

            \vspace{-2mm}
            \begin{tabular}{lcccc}
                \toprule
                 & Win\% & Draw\% & Lose\%\\  
                \midrule 
                Image Alignment  & 87.28 & 3.28 & 9.42 \\
                3D Appearance  &  77.63 &6.14 & 16.22\\
                Motion Quality & 70.17 & 6.57 & 23.24\\
                \bottomrule
                \label{tab:user_study}
            \end{tabular}
        \end{center}
    \end{minipage}
    \hfill
    \begin{minipage}[t]{0.53\textwidth}
        \raggedright
        \centering
        \vspace{-4mm}
        \includegraphics[width=\textwidth]{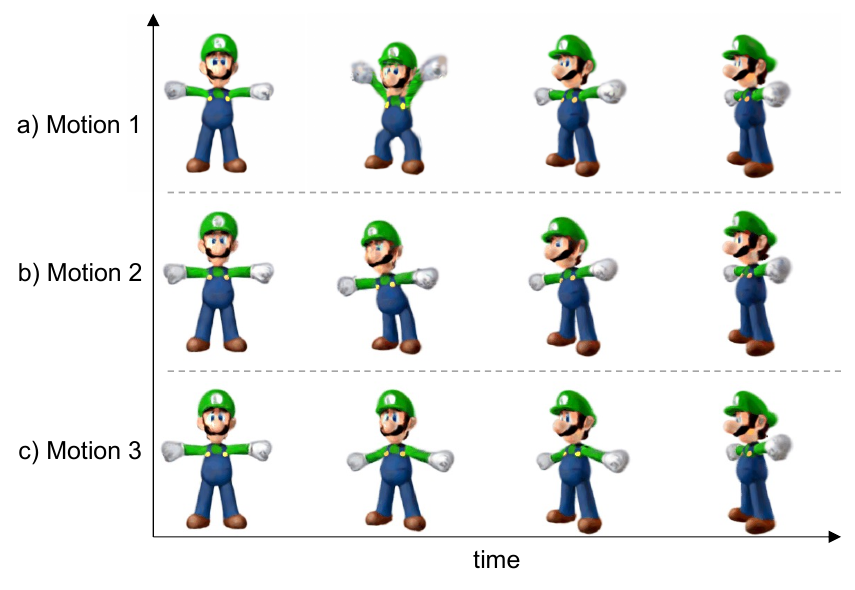}
        \vspace{-6.5mm}
        \caption{\textbf{Controllable Motions.} DreamGaussian4D allows easy control of the generated motions. Different 3D motions can be generated from different driving videos for the same input image.}
        \label{fig:ablation_motion}
    \end{minipage}
\end{figure}

\textbf{Diverse Motions.}
Different from most existing 4D generation approaches~\cite{bahmani20234d,jiang2023consistent4d} using SDS, our method allows better controllability and more diversity in the motions.
Different 4D motions can be generated from different driving videos. In Fig.~\ref{fig:ablation_motion}, we generate three different driving videos for an input image, which results in three distinct 3D motions.

\textbf{Dynamic Scene Composition.}
By exporting 4D GS to textured meshes, we could directly generate composited dynamic scenes with engines (e.g., Blender).
The rendered example images from different view angles could be visualized in Fig.~\ref{fig:teaser}.
More qualitative results can be found in the appendix and our project page.

\textbf{4D Generation based on LGM.}
LGM~\cite{tang2024lgm} is a recently proposed feedforward method for image-to-3D generation using 3DGS, which offers faster generation with higher quality compared to SDS-based methods. 
By optimizing virtual camera settings, we facilitate 4D GS generation from the static 3D output of LGM.
Qualitative 4D generation results are illustrated in Fig.~\ref{fig:lgm4d}.

\section{Conclusion}\label{sect:conclusion}
We propose {\OM{} (\OMS{})}, an efficient image-to-4D generation framework
with 4D Gaussian Splatting. 
\OMS{} significantly reduces the optimization time from several hours to minutes. 
Moreover, we show that driving motion generation using generated videos allows explicit control of the 3D motion.
Lastly, \OMS{} allows mesh extraction and temporally coherent texture optimization, which facilitates real-world applications.

\textbf{Limitations.}
Despite high efficiency, the generation quality is not as high as many photo-realistic 4D generation methods. 
In addition, we assume a good-quality driving video, which would largely influence the 4D generation quality.

\textbf{Broader Impacts.} The approach enables fast generation of dynamic 3D content and can be widely applied to videos, gaming, and mixed reality applications. However, it should be used with caution to prevent malicious impersonation.

\bibliographystyle{plain}
\bibliography{ref}

\appendix
\newpage
\appendix
\onecolumn

\section{More implementation details}\label{sect:implementation}
\subsection{3D Gaussian Splatting settings.}
We initialized the static 3D GS with 5,000 Gaussians. We did not use the Spherical Harmonics. For DreamGaussianHD, we set the dense percentage to 0.1, the densification interval to 100, densification gradient threshold to 0.05. All other settings follow DreamGaussian~\cite{tang2023dreamgaussian}.

\subsection{HexPlane settings}
We use 32 for both spatial and temporal resolutions. No multi-resolution is applied. The grid feature dimension is 32. We set the learning rate for the MLP to 0.00064 and the learning rate for the HexPlane grid to 0.0064. Other settings follow 4D Gaussian Splatting~\cite{wu20234d}.

\subsection{Evaluation settings.}
\paragraph{Image-to-4D evaluation.} For the user study, we describe the evaluation axis to participants as follows:
\begin{itemize}
    \item Image Alignment. Which one looks the most like the input image?
\item 3D Appearance. Which one has a better-defined 3D geometry and a nicer look?
\item Motion Quality. Which one has a more natural and realistic motion?
\end{itemize}
For the computation of the CLIP-I metric, we use the 8 images provided on their official website.
\paragraph{Video-to-4D evaluation.} We use the code and data provided by~\cite{jiang2023consistent4d} in their officially released code. All videos are rendered from animated character assets. A horizontal angle of 0 degrees is used for input and other novel views are evaluated at -75 degrees, 15 degrees, 105 degrees, and 195 degrees. Similarity metric can be thus computed between the generated 4D content to the original animated character.
\subsection{4D Generation based on LGM}
We can use LGM~\cite{tang2024lgm} in replace of DreamGaussianHD to generate static 3D from the input image or a frame in the input video. Since LGM often generates 3D GS that misaligns with the input image, we render the LGM model from [-180$^\circ$, 180 $^\circ$] azimuths at 0$^\circ$ elevation, and select the azimuth that has the lowest L2 distance to the input image on the RGB space. Then, we render the 3D GS from the selected azimuth and the other three orthogonal viewpoints to get four multiview images. Finally, we feed the multiview images into LGM to get the static initialization that aligns well with the input image. Following LGM's camera setting, we use a camera radius of 1.5 in the dynamic optimization.

\section{More quantitative results}
We show detailed evaluation metrics for the video-to-4D benchmark in~\autoref{tab:con4d}.
\begin{table*}[t!] %
\centering
\caption{Quantitative comparisons between ours and others on Consistent4D dataset. Baseline results are from~\cite{gao2024gaussianflow}.
}
\label{tab:con4d}
\scalebox{0.6}
{\begin{tabular}{lcccccccccccccccc}
\toprule
\multirow{2}[2]{*}{Method}  & \multicolumn{2}{c}{\textbf{Pistol}} & \multicolumn{2}{c}{\textbf{Guppie}}& \multicolumn{2}{c}{\textbf{Crocodile}}& \multicolumn{2}{c}{\textbf{Monster}}& \multicolumn{2}{c}{\textbf{Skull}} & \multicolumn{2}{c}{\textbf{Trump}} & \multicolumn{2}{c}{\textbf{Aurorus}} & \multicolumn{2}{c}{\textbf{Mean}} \\ \cmidrule(lr){2-3} \cmidrule(lr){4-5} \cmidrule(lr){6-7} \cmidrule(lr){8-9} \cmidrule(lr){10-11} \cmidrule(lr){12-13}\cmidrule(lr){14-15}\cmidrule(lr){16-17}
 &  LPIPS$\downarrow$  & CLIP$\uparrow$  & LPIPS$\downarrow$  & CLIP$\uparrow$ & LPIPS$\downarrow$  & CLIP$\uparrow$ & LPIPS$\downarrow$  & CLIP$\uparrow$ & LPIPS$\downarrow$  & CLIP$\uparrow$ & LPIPS$\downarrow$  & CLIP$\uparrow$ & LPIPS$\downarrow$  & CLIP$\uparrow$  & LPIPS$\downarrow$ &CLIP$\uparrow$\\
\midrule
D-NeRF~\cite{pumarola2021d}  &  0.52 & 0.66 & 0.32 & 0.76 & 0.54 & 0.61 & 0.52 & 0.79 & 0.53 & 0.72 & 0.55 & 0.60 & 0.56 & 0.66 & 0.51 & 0.68  \\
K-planes~\cite{fridovich2023k}  & 0.40 & 0.74 & 0.29 & 0.75 & 0.19 & 0.75 & 0.47 & 0.73 & 0.41 & 0.72 &   0.51 & 0.66 & 0.37 & 0.67 & 0.38 & 0.72 \\
C4D~\cite{jiang2023consistent4d}  & 0.10 & 0.90 & 0.12 & 0.90 & 0.12 & 0.82 & 0.18 & 0.90 & 0.17 & 0.88 & 0.23 & 0.85 & 0.17 & 0.85 & 0.16 & 0.87 \\
GauFlow~\cite{gao2024gaussianflow}  &  0.10 & 0.94 & \textbf{0.10} & \textbf{0.93} &  \textbf{0.10} & \textbf{0.90} & 0.17 & 0.92 & 0.17 & 0.92 & 0.20 & 0.85 & 0.15 & \textbf{0.89} & 0.14 & 0.91 \\ 
\midrule
Ours  & \textbf{0.08} & \textbf{0.94} & \textbf{0.10}  & \textbf{0.94}  &  \textbf{0.10} & 0.89 & \textbf{0.15} & \textbf{0.95} & \textbf{0.13} & \textbf{0.95} & \textbf{0.16} & \textbf{0.90} & \textbf{0.13} & 0.88 & \textbf{0.12} & \textbf{0.92}\\
\bottomrule
\end{tabular}}
\vspace{-10pt}
\end{table*}

\section{More qualitative results}
\begin{figure*}[ht!]
\begin{center}
\includegraphics[width=\textwidth]{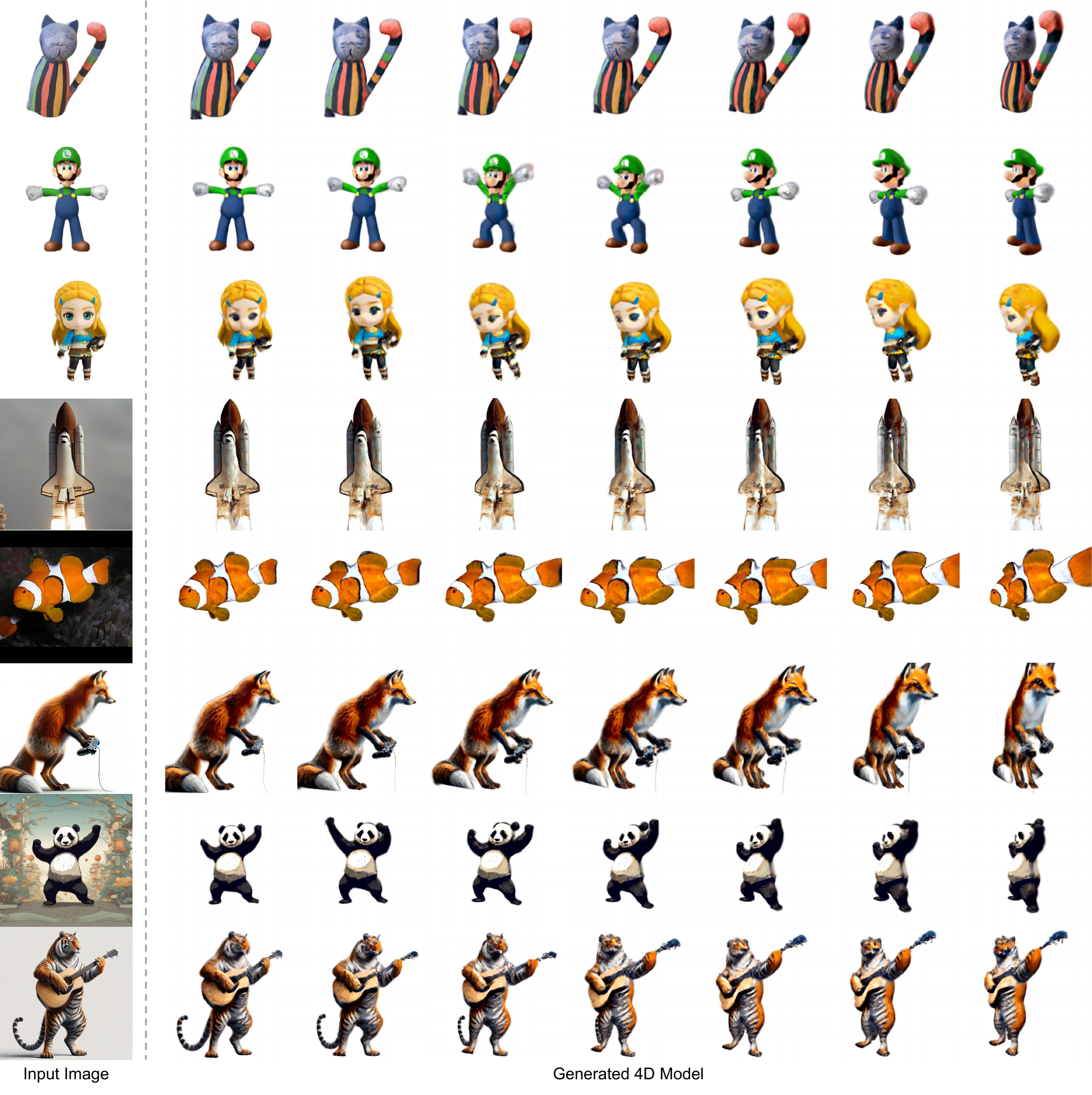}
\caption{\textbf{Qualitative results.} Renders are shown with gradually changing time stamps and view angles.}
\label{fig:qualitative_full}
\end{center}
\end{figure*}

More image-to-4D results are provided in ~\autoref{fig:qualitative_full}.

\subsection{More qualitative result on video to 4D}
\begin{figure*}[ht!]
\begin{center}
\includegraphics[width=0.8\textwidth]{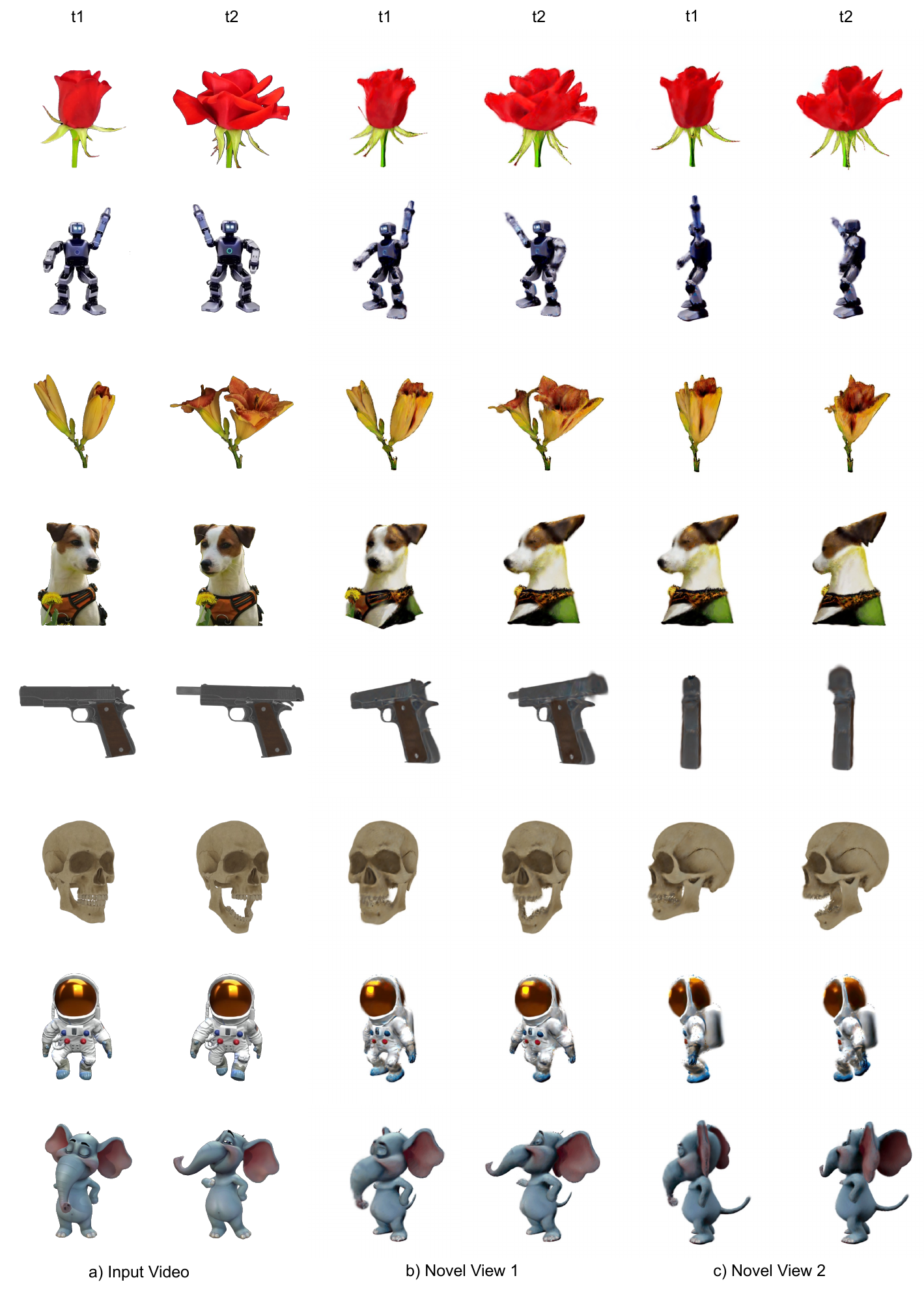}
\caption{\textbf{Qualitative results for video-to-4D.} Renders are shown in two novel views at two timesteps.}
\label{fig:qualitative_v24d_full}
\end{center}
\end{figure*}

More video-to-4D results are provided in ~\autoref{fig:qualitative_v24d_full}.

\section{More Ablations and Analysis}
\subsection{Deformation field representations.}
\begin{figure}[ht!]
\begin{center}
\includegraphics[width=\columnwidth]{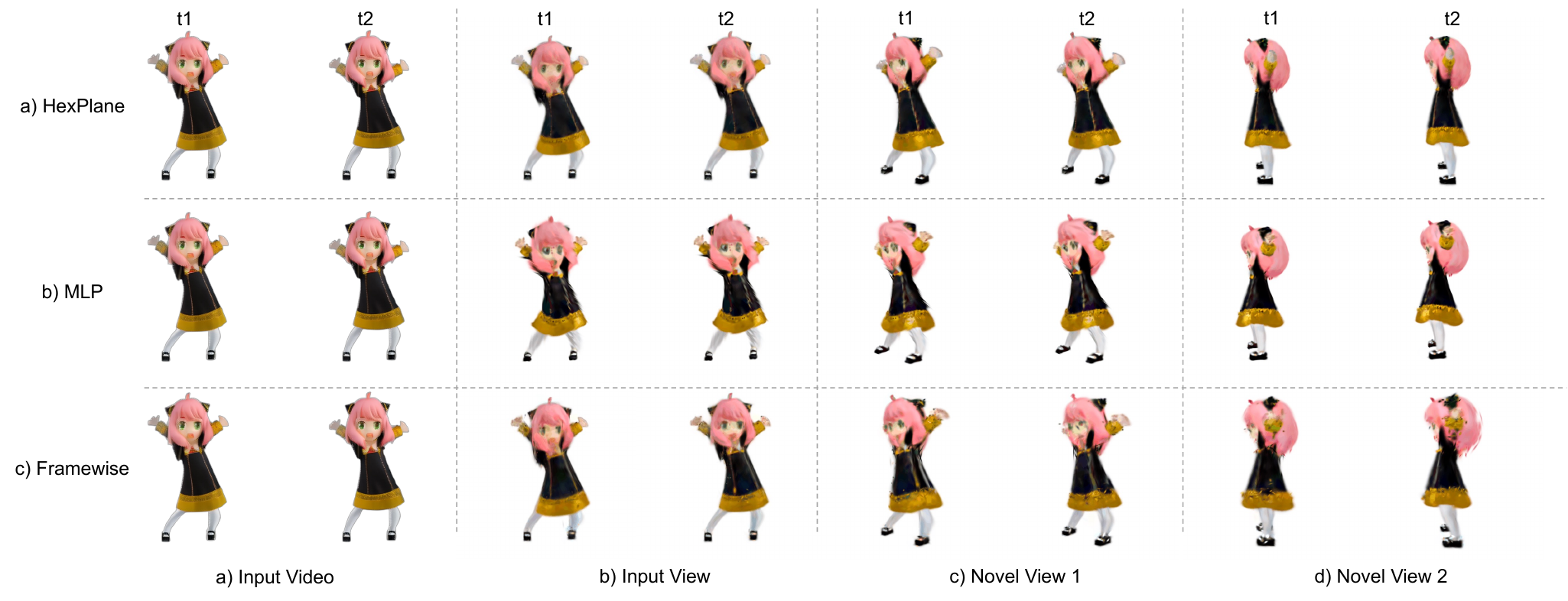}
\caption{\textbf{Ablation on motion representation.} }
\label{fig:ablation_motion_representation}
\end{center}
\end{figure}

We analyze the effect of different deformation field representations, including a pure MLP network and frame-wise Gaussian. The pure MLP network takes the position embedding and time embedding as the input, and outputs deformation variables. The frame-wise Gaussians use the static 3D GS as an initialization and fit a separate 3D GS for each video frame independently. We train them under the same setting and show the result in ~\autoref{fig:ablation_motion_representation}. The MLP network is under-optimized and the frame-wise Gaussian lacks temporal consistency.

\subsection{Temporal loss.}
\begin{figure}[ht!]
\begin{center}
\includegraphics[width=0.6\columnwidth]{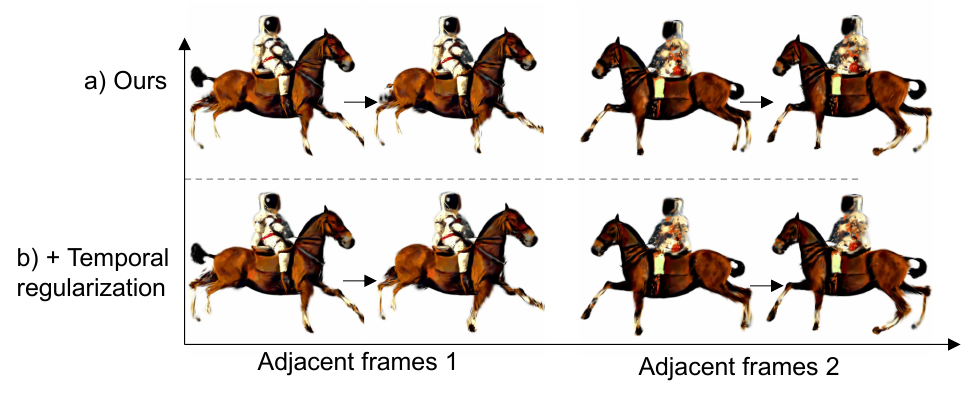}
\caption{\textbf{Ablation on temporal loss.} }
\label{fig:ablation_temporal_loss}
\end{center}
\end{figure}

We add the temporal loss described by ~\cite{luiten2023dynamic} to the optimization but can not observe significant improvement. A brief comparison is shown in \autoref{fig:ablation_temporal_loss}.

\end{document}